\documentclass{esannV2}
\usepackage{graphicx}
 \usepackage{url}
\usepackage{xcolor}

\usepackage[latin1]{inputenc}
\usepackage{amssymb,amsmath,array}
\usepackage{comment}

\begin{document}
\title{Deviations from the NTK Regime in CNNs\\ under Joint Depth-Width Scaling}

\author{Maria Matveev$^1$, Pascal Esser$^1$, Gitta Kutyniok$^{1}$, Mariia Seleznova$^2$
\thanks{GK is further affiliated with University of Troms{\o} and DLR-German Aerospace Center. MM and GK are supported by the DAAD programme Konrad Zuse Schools of Excellence in Artificial Intelligence, sponsored by the Federal Ministry of Research, Technology and Space.
MS is funded by the Deutsche Forschungsgemeinschaft (DFG, German Research Foundation) --- Project number 572957656 (``RMT4DL: Random Matrix Theory for Deep Learning'').}
\vspace{.3cm}\\
1- LMU Munich -- Department of Mathematics\\
 Munich Center for Machine Learning 
\vspace{.1cm}\\
2- TU Darmstadt -- Signal Theory and Statistical Learning Group
}

\maketitle

\begin{abstract}
The concentration and training-time constancy of the neural tangent kernel (NTK) depend on the interplay between network width, depth, and initialization. While joint depth-width scaling has been studied in fully-connected, residual, and recurrent networks, its effects in convolutional neural networks (CNNs) remain less understood. We investigate these effects numerically in CNNs, examining NTK dispersion at initialization and kernel rotation during training. Let depth $L$ be the number of convolutional layers, and width $C$ the number of channels per layer. We find that NTK dispersion increases with $L/C$ in the critical and chaotic initialization regimes, but remains small in the ordered regime. At critical initialization, dispersion curves approximately collapse as a function of $L/C$, suggesting a nontrivial joint scaling regime. During training, kernel rotation generally increases with $L/C$, even in the ordered regime where initial dispersion is small. Together, these results suggest that $L/C$ may govern NTK deviations from the infinite-width limit in certain regimes, and that NTK concentration at initialization and stability during training exhibit different scaling behaviors.
\end{abstract}

\section{Introduction}

The \emph{Neural Tangent Kernel (NTK)} provides a compact description of the training dynamics of neural networks under gradient descent. For a scalar-valued network $f_\theta$, the NTK is defined as
$    \Theta_\theta(x,x'):=
\left\langle
\nabla_\theta f_\theta(x),
\nabla_\theta f_\theta(x')
\right\rangle
$, where $\theta$ denotes the network's trainable parameters and $x,x'$ are data points. In the infinite-width limit at fixed depth, under suitable assumptions, the NTK \emph{becomes deterministic} at initialization and \emph{remains constant during training} \cite{jacot2020ntk, lee19wide, arora2019cnnntk}. This gives rise to the so-called \emph{NTK regime}, in which training is governed by a fixed kernel and becomes analytically tractable.
However, these results do not establish whether the NTK regime persists when \emph{width and depth scale jointly}.

Prior work on multilayer perceptrons (MLPs) has shown that finite-width effects can accumulate when depth grows proportionally with width, making the \emph{depth-to-width ratio} a key scaling parameter governing deviations from the infinite-width limit \cite{hanin2020finite,seleznova2022neural,hanin2024random,roberts2022principles}. The effect of this ratio further depends on initialization, with qualitatively different behavior in the \emph{ordered} and \emph{chaotic} regimes and at their critical boundary, the \emph{edge of chaos} (EOC) \cite{poole2016regimes,xiao2020disentangling,seleznova2022analyzing,hayou2019exact}. Thus, both depth-to-width scaling and initialization determine the applicability of the NTK regime in deep MLPs.

Importantly, deviations from the infinite-width limit are architecture-depen\-dent. Unlike in MLPs, depth and width limits commute in residual networks under suitable residual-branch scaling, with no nontrivial joint scaling limit arising~\cite{hayou2023width,bordelon2024depthwise,yang2024tensor}. In contrast, finite-width effects in linear recurrent networks accumulate more rapidly with depth, giving rise to a distinct joint scaling regime where depth grows proportionally to the square root of width \cite{seleznova2026long,bar2026revisiting}.

These results motivate studying depth-width scaling in CNNs, where spatial weight sharing introduces structure absent from the architectures discussed above and may lead to new depth-width scaling effects. In CNNs, the natural width parameter is the number of channels $C$, which suggests the ratio $L/C$ as a candidate scaling parameter by analogy with MLPs. In this work, we investigate empirically whether this ratio governs NTK concentration at initialization and stability during training. Additionally, a direct comparison of the behavior of MLP and CNN is presented in App.~\ref{app:MLP}.\!\footnote{Code is available at \url{https://github.com/Maascha/cnn_ntk}.}

\section{Experimental Setup}

\paragraph{CNN architecture.}
We consider CNNs with $L$ convolutional layers of $C$ channels each, using $\tanh$ activations, stride one, and circular padding. Let $C_\ell$ denote the number of channels at layer $\ell$, with $C_0=3$ and $C_\ell=C$ for $\ell\geq1$. Given an input $x^0=x\in\mathbb R^{C_0\times H\times W}$, the activations at layer $\ell\in[L]$ are
\begin{equation}\label{eq:activation}
x^\ell_{\beta,p}
=\tanh\left(
\sum_{\alpha=1}^{C_{\ell-1}}
\sum_{u\in\mathcal K}
W^\ell_{\beta\alpha u},x^{\ell-1}_{\alpha,p+u}
+b^\ell_\beta
\right),
\end{equation}
where $\alpha,\beta$ index channels, $p$ denotes a spatial position on the $H\times W$ grid, and $\mathcal K$ is the set of filter offsets with $|\mathcal K|=k^2$. Spatial indices are taken modulo the grid dimensions, set to $H=W=16$ in our experiments. The scalar output is produced by global average pooling followed by a linear readout, namely $f_\theta(x)=\sum_{\beta=1}^{C}w_\beta(HW)^{-1}\sum_{p}x^L_{\beta,p}+b$, where $w_\beta,b$ are the readout parameters.

\paragraph{Parametrization and initialization.}
We use Standard Parametrization (SP)\footnote{Up to learning rate rescaling, both NTK Parametrization (NTP) and Standard Parametrization (SP) lead to the kernel regime in the infinite-width limit \cite{yang2024tensor,lee19wide}.}, initializing all parameters independently as follows:
$$
W^\ell_{\beta\alpha u}\sim\mathcal N\left(0,\frac{\sigma_w^2}{C_{\ell-1}k^2}\right),
\quad
w_\beta\sim\mathcal N\left(0,\frac{\sigma_w^2}{C}\right), \quad 
b^\ell_\beta, b \sim \mathcal{N}(0,1).
$$
To examine how initialization affects depth-width scaling, we conduct experiments in three regimes: ordered ($\sigma_w^2=1$), critical ($\sigma_w^2=3.44$, at the EOC for $\sigma_b^2=1$), and chaotic ($\sigma_w^2=5$) \cite{poole2016regimes}.

\paragraph{Data and training.}
We use a binary CIFAR-10 task (airplane vs.\ frog) \cite{Krizhevsky2009LearningML}, with $2048$ class-balanced training images and $2000$ test images. Images are downsampled to $16\times16$ by average pooling, centered using the training-set pixel mean, and rescaled to norm $\sqrt{768}$. We train the centered network $f_\theta-f_{\theta_0}$ on the summed squared error $\mathcal L(\theta)=\tfrac12\sum_{i=1}^{n}(f_\theta(x_i)-f_{\theta_0}(x_i)-y_i)^2$ by full-batch gradient descent, without momentum or weight decay. For each initialization, the learning rate is set to $\eta=1/\sigma_{\max}(\Theta_0(X,X))$, where $X$ is the full training set, $\Theta_0$ is the NTK at initialization and $\sigma_{\max}(\cdot)$ denotes the largest eigenvalue. Training stops when the mean squared error reaches $0.1$ or after $10^5$ steps. We use three independent training seeds per configuration.

\section{NTK Variability at Initialization}

\begin{figure}[h!]
    \centering
    \includegraphics[height=0.24\textwidth,trim={0 0.1cm 0 0.1cm},clip]{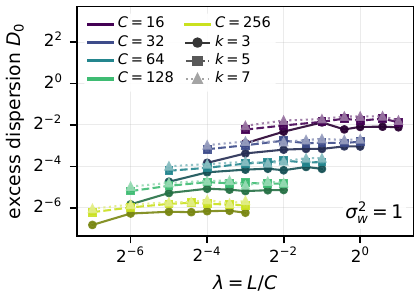}\hfill
    \includegraphics[height=0.24\textwidth, trim={16pt 0.1cm 0 0.1cm}, clip]{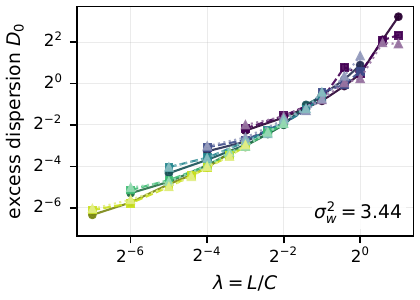}\hfill
    \includegraphics[height=0.24\textwidth, trim={16pt 0.1cm 0 0.1cm}, clip]{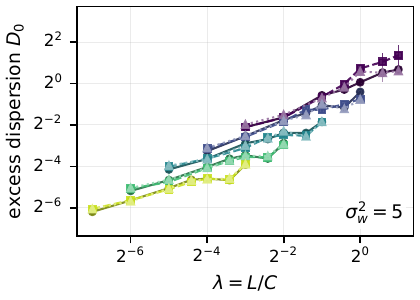}
    \caption{\textbf{NTK dispersion $D_0$ versus depth-to-width ratio $\lambda:=L/C$.} Each point averages $D_0(x)$ over $16$ images, with dispersion estimated from 1000 independent initializations per image. Standard errors are smaller than the markers except at the largest $\lambda$. We vary channel count $C$, filter size $k$, and initialization variance $\sigma_w^2$.}
    \label{fig:init}
\end{figure}

We quantify NTK fluctuations at initialization by the \emph{excess dispersion} of its diagonal entries, defined for a fixed input $x$ as
$$
D_0(x):=\frac{\mathbb E[\Theta_0(x,x)^2]}
{\mathbb E[\Theta_0(x,x)]^2}-1,
$$
where the expectation is over independent initializations. This quantity vanishes at infinite width, when the NTK becomes deterministic. Fig.~\ref{fig:init} shows the empirical dependence of $D_0$ on the depth-to-width ratio $L/C$ across the three initialization regimes.

Fig.~\ref{fig:init} reveals two main trends. First, NTK dispersion increases with the depth-to-width ratio $L/C$ in the critical ($\sigma_w^2=3.44$) and chaotic ($\sigma_w^2=5$) regimes, but not in the ordered regime ($\sigma_w^2=1$). This mirrors the qualitative behavior observed in MLPs \cite{seleznova2022analyzing,seleznova2022neural}.

Second, at critical initialization, curves for different $C$ approximately collapse at sufficiently large $L$ and $C$, suggesting a nontrivial joint depth-width scaling regime governed by the ratio $L/C$. In contrast, the curves remain separated across widths in the ordered and chaotic regimes, indicating that $L/C$ alone does not fully capture NTK dispersion and that additional dependence on $C$ persists.

\section{NTK Behavior During Training}

To quantify changes in the NTK during training, we measure its \emph{rotation} relative to initialization,
$$
R_t:=1-\frac{\langle\Theta_t,\Theta_0\rangle_F}
{\|\Theta_t\|_F\|\Theta_0\|_F},
$$
where $\Theta_t=\Theta_{\theta_t}(X_P,X_P)$ is the NTK matrix at training step $t$, evaluated on a fixed, class-balanced set $X_P$ of $P=256$ training images. Here $\langle\cdot,\cdot\rangle_F$ and $\|\cdot\|_F$ denote the Frobenius inner product and norm. This metric is invariant to overall kernel rescaling and vanishes when the kernel remains constant.

\begin{figure}[h!]
    \centering
      \includegraphics[height=0.24\textwidth,trim={0 0.1cm 0 0.1cm}, clip]{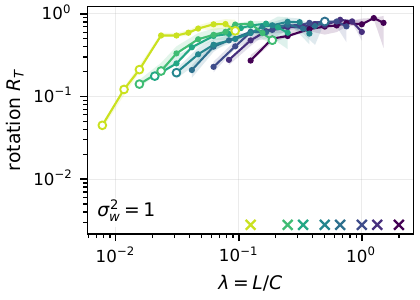}\hfill
    \includegraphics[height=0.24\textwidth, trim={16pt 0.1cm 0 0.1cm}, clip]{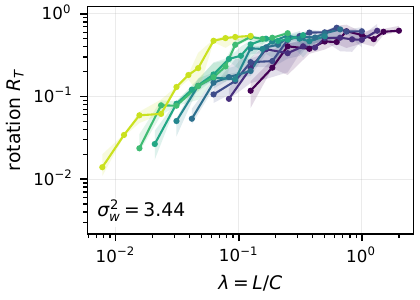}\hfill
    \includegraphics[height=0.24\textwidth, trim={16pt 0.1cm 0 0.1cm}, clip]{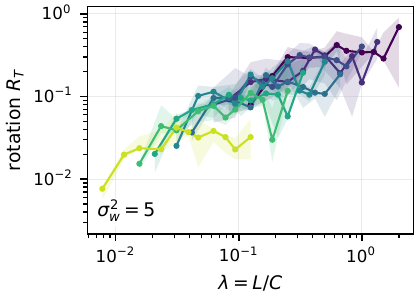}
    \includegraphics[width=\textwidth, trim={0 .3cm 0 .3cm}, clip]{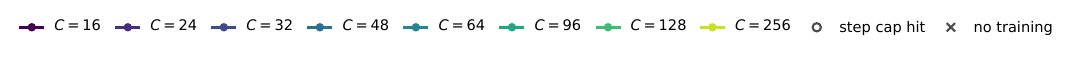}
    \caption{\textbf{Final NTK rotation $R_T$ versus depth-to-width ratio $\lambda$}, for $3\times3$ filters.  Training stops at step $T$ at $\mathrm{MSE}=0.1$ or the iteration limit. Lines show the median over three seeds and bands
the minimum--maximum range. Open markers mean at least one seed reached the iteration limit, in which case $R_T$ is evaluated there. Crosses mark configurations in which no seed trained, defined as training MSE above $0.99$ at the iteration limit for all seeds.}
    \label{fig:training}
\end{figure}
Fig.~\ref{fig:training} shows the final NTK rotation $R_T$ against the depth-to-width ratio $\lambda$ for $3\times3$ filters. Overall, $R_T$ increases with $\lambda$, indicating that the NTK changes more substantially during training as depth grows relative to width. Notably, this trend is also present in the ordered regime ($\sigma_w^2=1$), where the NTK has small dispersion at initialization yet undergoes substantial rotation during training. An additional factor to consider is the poor trainability of deep networks in this regime \cite{xiao2020disentangling}: reaching the target error may require more gradient descent steps, potentially contributing to greater kernel rotation. Thus, NTK concentration at initialization does not necessarily imply stability during training.

At EOC initialization ($\sigma_w^2=3.44$), the rotation curves separate across channel counts $C$, despite the approximate collapse observed for initial dispersion. This suggests that $\lambda$ alone does not characterize training-induced NTK changes. Interestingly, the curves at $\sigma_w^2=5$ align more closely, providing preliminary evidence that kernel rotation may be governed by $\lambda$ in the chaotic regime.

\section{Initial NTK Dispersion vs. Training-Induced Rotation}
\begin{figure}[h!]
    \centering
  \includegraphics[height=0.21\textwidth,trim={0 0.1cm 0 0.1cm},clip]{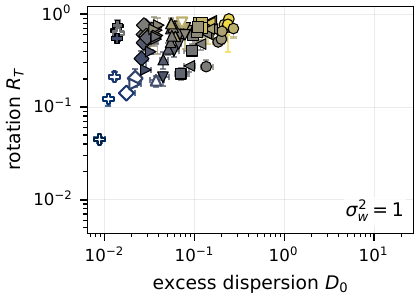}\hfill
    \includegraphics[height=0.21\textwidth, trim={16pt 0.1cm 0 0.1cm}, clip]{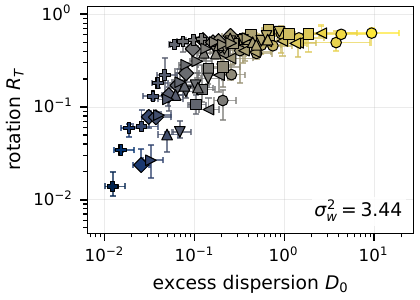}\hfill
      \includegraphics[height=0.21\textwidth, trim={16pt 0.1cm 0 0.1cm}, clip]{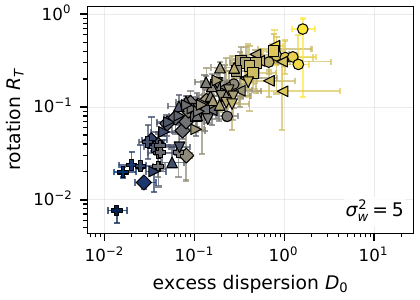}\hfill
    \includegraphics[height=0.21\textwidth]{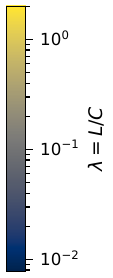} 
    \includegraphics[width=.8\textwidth,trim={0 .3cm 0 .3cm}, clip]{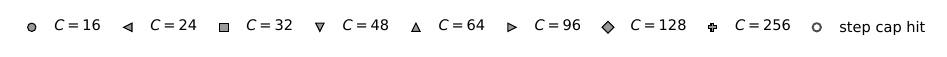}
    \caption{\textbf{Final NTK rotation $R_T$ versus initial dispersion $D_0$.} Rotation is summarized by the median across training seeds, while dispersion is averaged over 16 inputs. Error bars indicate the corresponding minimum--maximum ranges. Colors encode $\lambda$, and markers encode $C$.}
    \label{fig:initvstrain}
\end{figure}

Motivated by the different scaling behavior of initial NTK dispersion and training-induced rotation, we examine the correlation between these quantities across network configurations in Fig.~\ref{fig:initvstrain}. In the ordered regime ($\sigma_w^2=1$), the correlation is only moderate (Spearman $\rho=0.45$): at fixed width, rotation can increase substantially with depth while dispersion remains nearly constant. For instance, at $C=256$, rotation grows by an order of magnitude with little change in dispersion. At critical and chaotic initialization, the two quantities are more strongly correlated across the tested architectures.

\section{Discussion and Future Work}

This work presents a preliminary numerical study of joint depth-width scaling in CNNs, focusing on NTK variability at initialization and its evolution during training. Our results show that increasing depth generally amplifies deviations from infinite-width NTK behavior. The depth-to-width ratio $L/C$ appears to capture these effects in certain settings, notably for the initial NTK dispersion at critical initialization, but not universally across initialization regimes. 

Our findings are currently limited to a specific CNN family, one binary classification task, and the observed finite-size scaling does not establish an asymptotic joint limit. Therefore, beyond exploring further architectures, a natural direction for future work is to explain these observations theoretically by identifying nontrivial joint depth-width limits in CNNs, potentially governed by $L/C$. More broadly, this work contributes to understanding joint scaling regimes across architectures beyond MLPs and the deviations from infinite-width descriptions of neural networks, particularly the NTK regime.
\clearpage

\begin{footnotesize}

\bibliographystyle{unsrt}
\bibliography{references}

\end{footnotesize}

     \appendix
     \clearpage

\section{Comparison with Multilayer Perceptrons}
\label{app:MLP}

To directly illustrate the comparison in the behavior of MLPs and CNNs, we repeat the experimental settings from CNNs from the main paper using multilayer perceptrons (MLPs). To match MLP settings to those of CNNs, we flatten the input images into vectors $x\in\mathbb R^{768}$, use $\tanh$ activations, a linear readout, and hidden-layer width $M=9C$, where $C$ is the number of CNN channels. Note that this choice of width matches the hidden-layer fan-in dimension of CNNs with $3\times3$ filters: each convolutional preactivation sums over $9C$ input-channel and filter-offset combinations (see Eq.~\eqref{eq:activation}). We therefore plot against $\lambda=9L/M=L/C$, matching the depth-to-fan-in ratio across architectures.

\begin{figure}[h!]
    \centering
      \includegraphics[height=0.24\textwidth,trim={0 0.1cm 0 0.1cm}, clip]{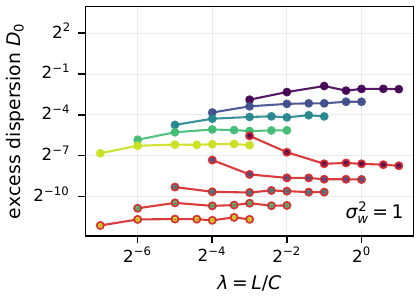}\hfill
    \includegraphics[height=0.24\textwidth, trim={16pt 0.1cm 0 0.1cm}, clip]{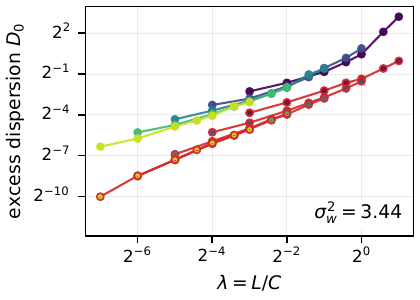}\hfill
    \includegraphics[height=0.24\textwidth, trim={16pt 0.1cm 0 0.1cm}, clip]{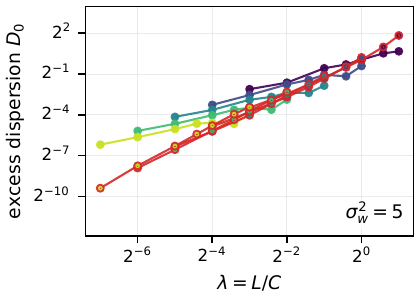}
    \includegraphics[width=\textwidth, trim={0 .3cm 0 .3cm}, clip]{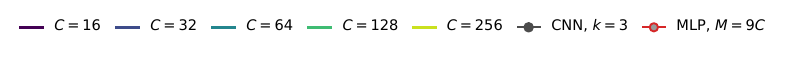}
    \caption{\textbf{NTK dispersion $D_0$ versus depth-to-width ratio $\lambda:=L/C=9L/M$ for the CNN and MLP.} Each point averages $D_0(x)$ over $16$ images, with dispersion estimated from 1000 independent initializations per image. Standard errors are smaller than the markers except at the largest $\lambda$. We vary channel count $C$/hidden width $M=9C$, depth and initialization variance.}
    \label{fig:init_with_mlp}
\end{figure}

\begin{figure}[h!]
    \centering
      \includegraphics[height=0.24\textwidth,trim={0 0.1cm 0 0.1cm}, clip]{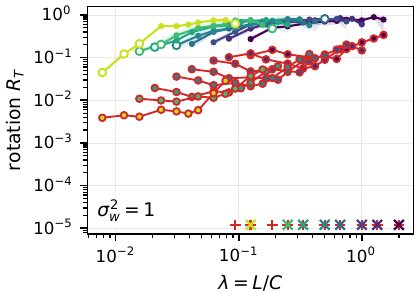}\hfill
    \includegraphics[height=0.24\textwidth, trim={16pt 0.1cm 0 0.1cm}, clip]{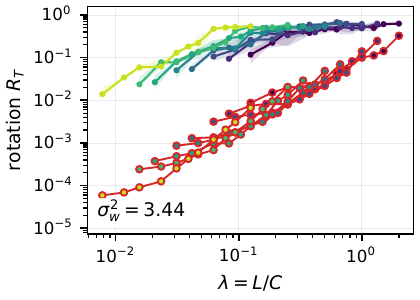}\hfill
    \includegraphics[height=0.24\textwidth, trim={16pt 0.1cm 0 0.1cm}, clip]{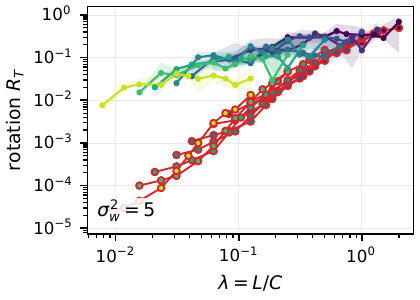}
    \includegraphics[width=\textwidth, trim={0 .3cm 0 .3cm}, clip]{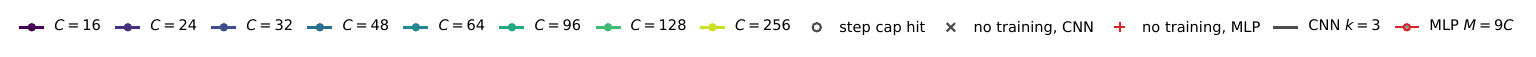}
    \caption{\textbf{Final NTK rotation $R_T$  versus depth-to-width ratio $\lambda$ for the MLP and CNN} with $3\times3$ filters.  Training stops at step $T$ at $\mathrm{MSE}=0.1$ or the iteration limit.}
    \label{fig:rotation_with_mlp}
\end{figure}

\begin{figure}[h!]
    \centering
      \includegraphics[height=0.21\textwidth,trim={0 0.1cm 0 0.1cm}, clip]{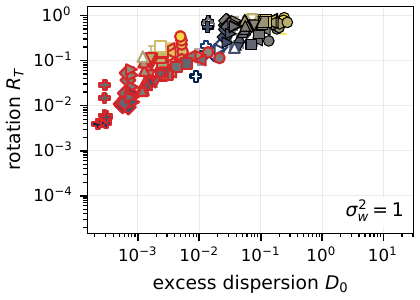}\hfill
    \includegraphics[height=0.21\textwidth, trim={16pt 0.1cm 0 0.1cm}, clip]{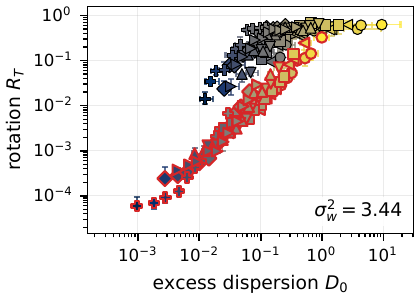}\hfill
    \includegraphics[height=0.21\textwidth, trim={16pt 0.1cm 0 0.1cm}, clip]{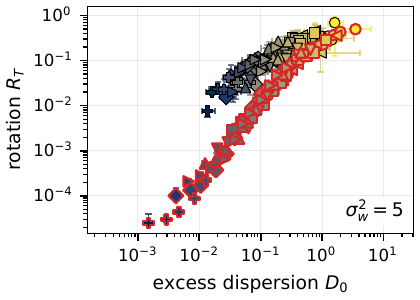}
        \includegraphics[height=0.21\textwidth]{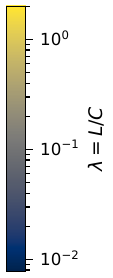} 
    \includegraphics[width=\textwidth, trim={0 0 0 .3cm}, clip]{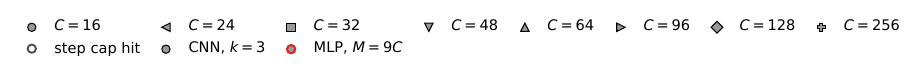}
    \caption{\textbf{Final NTK rotation $R_T$ versus initial dispersion $D_0$, for both MLP and CNN.} Rotation is summarized by the median across training seeds, while dispersion is averaged over $16$ inputs. Error bars indicate the corresponding minimum--maximum ranges. Colors encode $\lambda$, and markers encode $C$.}
    \label{fig:initvsrot_with_mlp}
\end{figure}

Figures~\ref{fig:init_with_mlp}, \ref{fig:rotation_with_mlp}, and \ref{fig:initvsrot_with_mlp} extend the Figures~\ref{fig:init}, \ref{fig:training}, and \ref{fig:initvstrain}  in the main text by including MLP results alongside CNN results.

\begin{figure}[th!]
    \centering
      \includegraphics[height=0.24\textwidth,trim={0 0.1cm 0 0.1cm}, clip]{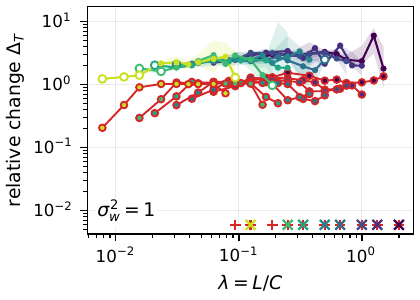}\hfill
    \includegraphics[height=0.24\textwidth, trim={16pt 0.1cm 0 0.1cm}, clip]{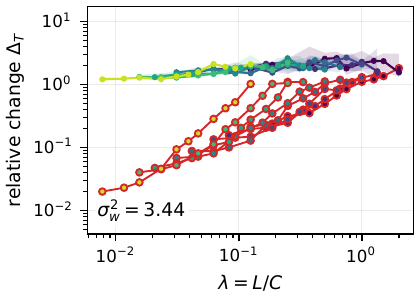}\hfill
    \includegraphics[height=0.24\textwidth, trim={16pt 0.1cm 0 0.1cm}, clip]{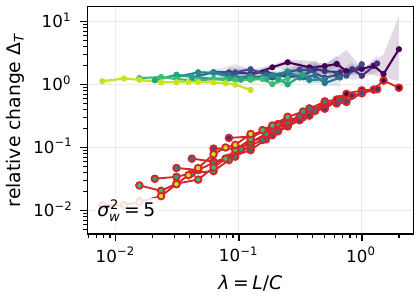}
    \includegraphics[width=\textwidth, trim={0 .3cm 0 .3cm}, clip]{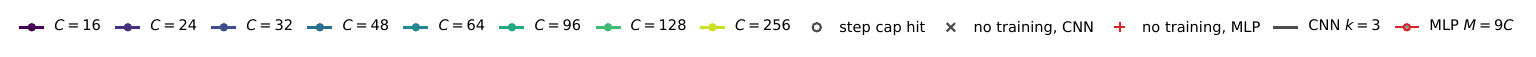}
    \caption{\textbf{Final Relative change $\Delta_T$ of the NTK versus depth-to-width ratio $\lambda$,  for both MLP and CNN}, with $3\times3$ filters.  Training stops at step $T$ at $\mathrm{MSE}=0.1$ or the iteration limit.}
    \label{fig:delta_with_mlp}
\end{figure}

\begin{figure}[h!]
    \centering
      \includegraphics[height=0.24\textwidth,trim={0 0.1cm 0 0.1cm}, clip]{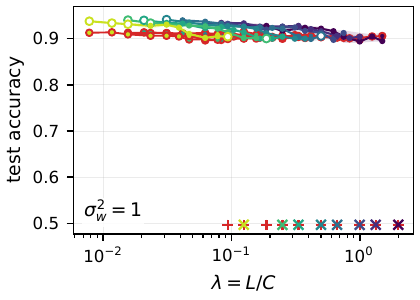}\hfill
    \includegraphics[height=0.24\textwidth, trim={16pt 0.1cm 0 0.1cm}, clip]{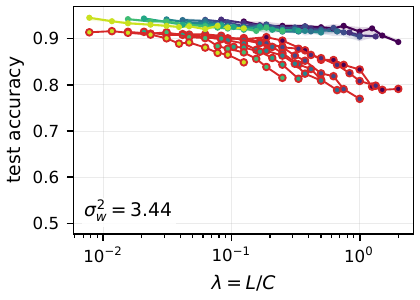}\hfill
    \includegraphics[height=0.24\textwidth, trim={16pt 0.1cm 0 0.1cm}, clip]{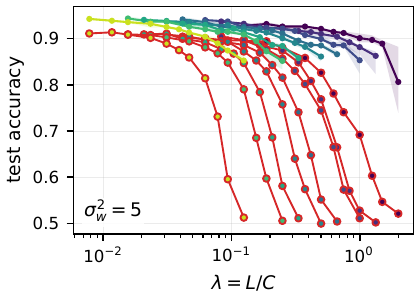} \includegraphics[width=\textwidth, trim={0 .3cm 0 .3cm}, clip]{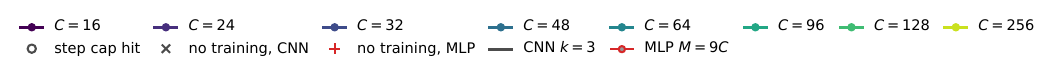}
    \caption{\textbf{Test accuracy at the end of training versus $\lambda$ for both MLP and CNN}, with $3\times3$ filters. }
    \label{fig:accuracy}
\end{figure}

\begin{figure}[h!]
    \centering
      \includegraphics[height=0.24\textwidth,trim={0 0.1cm 0 0.1cm}, clip]{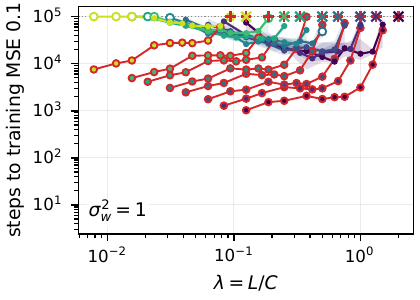}\hfill
    \includegraphics[height=0.24\textwidth, trim={16pt 0.1cm 0 0.1cm}, clip]{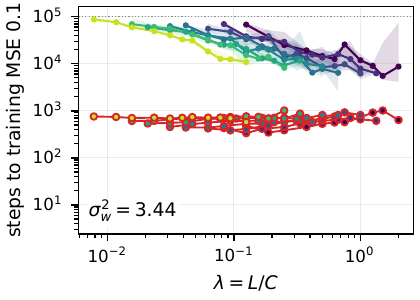}\hfill
    \includegraphics[height=0.24\textwidth, trim={16pt 0.1cm 0 0.1cm}, clip]{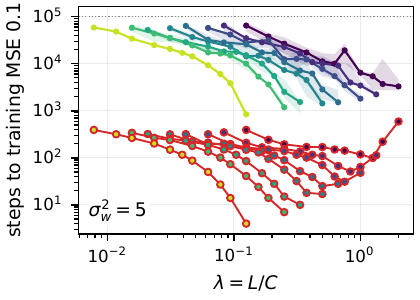}
    \includegraphics[width=\textwidth, trim={0 .3cm 0 .3cm}, clip]{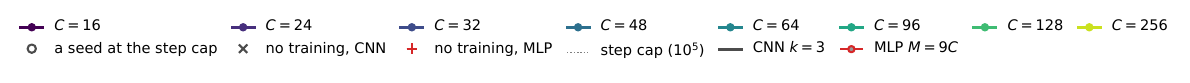}
    \caption{\textbf{Number of training steps to reach $\mathrm{MSE}=0.1$ for both MLP and CNN}, with $3\times3$ filters. The dotted line marks the iteration limit of $10^5$ steps.}
    \label{fig:steps}
\end{figure}

Furthermore, in addition to the kernel rotation $R_T$, we compare the relative change of the kernel during training in~Figure~\ref{fig:delta_with_mlp}. Figures \ref{fig:accuracy} and \ref{fig:steps} show the test accuracy at the end of training and the number of training steps for both architectures.

\paragraph{NTK variability at initialization.} Fig.~\ref{fig:init_with_mlp} compares the initial NTK excess dispersion $D_0$ of CNNs and MLPs. For MLPs, dispersion is governed by $L/M$ at the edge of chaos (EOC) and in the chaotic regime~\cite{seleznova2022neural}, consistent with the collapse of curves across widths $M$ in both regimes in the figure. CNNs exhibit similar behavior at the EOC, but their curves remain separated across channel counts $C$ in the chaotic regime, revealing an architecture-specific effect. Moreover, in the chaotic regime, the CNN dispersion grows slower with $L/C$ than the MLP dispersion with $9L/M$, whereas their growth rates appear comparable at the EOC. In the ordered regime, both architectures exhibit low dispersion with little dependence on the depth-to-width ratio.

\paragraph{NTK behavior during training.}
We train MLPs using the same protocol as CNNs and compare their final NTK rotation $R_T$ as a function of the depth-to-width ratio $\lambda$ in Fig.~\ref{fig:rotation_with_mlp}. In both architectures, rotation generally increases with $\lambda$ across all three initialization regimes. However, CNNs exhibit substantially larger rotation at small $\lambda$, indicating stronger deviations from the NTK regime, while MLP rotation appears to grow at a faster rate. Moreover, MLP rotation curves are more consistent with collapse across widths than their CNN counterparts, particularly at critical and chaotic initialization. In the ordered regime, the MLP curves remain separated at small $\lambda$ and converge only at large $\lambda$, near the depths where training fails to reach the target error.

For both CNNs and MLPs, Fig.~\ref{fig:delta_with_mlp} shows the \emph{relative change} of the NTK at the final step $T$,
$$
\Delta_T:=\frac{\|\Theta_T-\Theta_0\|_F}{\|\Theta_0\|_F}.
$$
In contrast to the rotation, $\Delta_T$ is not invariant to a rescaling of the kernel, i.e. it also measures a change of the kernel norm. For the CNNs at small $\lambda$, the rotation is small while the relative change is of order one, so the kernel approximately keeps its direction and changes mainly in norm. For the MLPs at critical and chaotic initialization, $\Delta_T$ becomes small as $\lambda$ decreases, so the kernel is nearly constant during training.

\paragraph{Final rotation versus initial dispersion.} Fig.~\ref{fig:initvsrot_with_mlp} adds the MLPs to Fig.~\ref{fig:initvstrain}. At critical and chaotic initialization, the MLP configurations fall close to a single curve, along which both quantities increase with $\lambda$. The CNN configurations lie above this curve, so that at the same initial dispersion the NTK of a CNN rotates considerably more than that of an MLP. In the ordered regime, the dispersion of an MLP is set by its width, while its rotation grows with depth, so that MLPs with nearly the same dispersion reach different rotations.

 \paragraph{Test accuracy.} Fig.~\ref{fig:accuracy} shows the test accuracy at the end of training. For CNNs, the test accuracy stays close to $0.9$ in all regimes and decreases only at the largest $\lambda$ in the chaotic regime. The test accuracy of the MLPs decreases with $\lambda$ at critical initialization and drops to chance level in the chaotic regime, i.e. at the same training error, the model does not generalize well. 

\paragraph{Training steps.} Fig.~\ref{fig:steps} shows the number of gradient descent steps needed to reach the target training error. At critical and chaotic initialization, all configurations reach the target error before the iteration limit. The number of steps decreases with $\lambda$ for the CNNs. In the ordered regime, the widest and shallowest CNNs stop at the iteration limit, and the deepest networks of both architectures do not train at all, consistent with the poor conditioning of the NTK in deep ordered networks~\cite{xiao2020disentangling}.

\section{Additional Experimental Details}
\label{app:details}

All the experiments are implemented in Python with PyTorch. The code is available at the GitHub repository: \url{https://github.com/Maascha/cnn_ntk}.

We vary the channel count $C\in\{16,24,32,48,64,96,128,256\}$ and depths $L\in\{2,3,4,6,8,10,12,16,20,24,32\}$. Each training run is repeated with three different seeds that change the network initialization. The data sets are the same for all runs. The networks are trained in single precision. The eigendecomposition of $\Theta_0$ that sets the learning rate for a setup and the computations of the dispersion use double precision.

The NTK is always computed with respect to all parameters, including the biases and the readout. During  training, we compute it using per-sample Jacobians, obtained through \texttt{torch.func}. At initialization,  the dispersion requires only the diagonal entries $\Theta_0(x,x)$, which we compute with NTK-vector products for $1000$ independent initializations per image. The initializations are drawn with the same random seed for every image. We estimate the ratio $\mathbb E[\Theta_0(x,x)^2]/\mathbb E[\Theta_0(x,x)]^2$ with the bias-corrected estimator of \cite[Appendix~C.3]{seleznova2022neural}.

We ran the experiments on heterogeneous compute hardware (NVIDIA TITAN RTX, RTX~3090, RTX~2080~Ti, RTX~4090, RTX~A6000 and H200). All experiments together required roughly $2{,}500$ GPU hours, more than half of which for the  CNNs in the ordered regime.

\end{document}